\documentclass{article} % For LaTeX2e
\usepackage[final]{colm2026_conference}

\usepackage{microtype}
\usepackage{hyperref}
\usepackage{url}
\usepackage{booktabs}
\usepackage{graphicx}


\usepackage{xcolor}

\usepackage{lineno}

\definecolor{darkblue}{rgb}{0, 0, 0.5}
\hypersetup{colorlinks=true, citecolor=darkblue, linkcolor=darkblue, urlcolor=darkblue}

\title{Legal Research Bench: Measuring End-to-End Reliability in Long-Horizon Legal Research Agents}

\makeatletter
\ifcolmfinal
\def\@maketitle{\vbox{\hsize\textwidth
{\Large\bf \@title\par}
{\centering
\def\And{\end{tabular}\hfil\linebreak[0]\hfil
        \begin{tabular}[t]{c}\bf\rule{\z@}{24pt}\ignorespaces}%
\def\AND{\end{tabular}\hfil\linebreak[4]\hfil
        \begin{tabular}[t]{c}\bf\rule{\z@}{24pt}\ignorespaces}%
\begin{tabular}[t]{c}\bf\rule{\z@}{24pt}\@author\end{tabular}\par}%
\vskip 0.3in minus 0.1in}}
\fi
\makeatother

\author{Katrina Drozdov\thanks{Corresponding author: katrina@vals.ai.}, Oliver Chen, Langston Nashold, Rayan Krishnan \\\\
Vals AI \\
San Francisco, USA \\
}

\begin{document}

\ifcolmsubmission
\linenumbers
\fi

\maketitle

% The final option sets the header to "Published as a conference paper at
% COLM 2026". This is a workshop paper, so override it after \maketitle.
\lhead{Published at the AI Measurement Science Workshop at COLM 2026.}

% =====================================================================
\begin{abstract}
    Legal research is a core and time-consuming legal workflow. Lawyers must identify controlling authority, verify that it remains valid, reconcile statutes and cases, and synthesize a grounded answer. Language model agents are a natural fit for this retrieval-intensive workflow, and automating even part of it would be valuable. But that value depends on reliability: a single missing authority, stale citation, or wrong legal conclusion can make an otherwise plausible answer unusable. We introduce \textbf{Legal Research Bench} (LRB), a benchmark of 413 open-ended U.S. legal research questions written by experts, each paired with a gold answer, supporting authorities, and a binary grading rubric. We evaluate thirteen frontier models in a harness with web search, case-law search, page parsing, and retrieval tools. We score agent responses through all-pass grading with source verification, where a response is correct only if every required criterion is satisfied and its cited authorities verify. We also validate the LLM judge against expert attorneys ensuring that benchmark scores track attorney judgment. Agents remain far from reliable: among the models we tested, the strongest, Claude Opus 4.8, is fully correct on 42.9\% of questions. Performance also varies substantially by task setting: all-pass rates differ across areas of law and are lower on questions requiring reconciliation of conflicting authorities. Across models, more turns, tool calls, and inference cost do not predict higher accuracy. 
\end{abstract}

% =====================================================================

\section{Introduction}
Legal research is one of the core tasks through which lawyers turn facts into legal advice, litigation strategy, compliance decisions, and client-facing work product. Doing it well requires locating controlling authority, verifying that it remains good law, reconciling conflicting statutes and cases, and synthesizing a grounded answer that withstands scrutiny. Errors are consequential: a missed exception, stale rule, or wrong holding can change the legal conclusion, so partial correctness does not translate to partial value.

This workflow is retrieval-intensive, iterative, and tool-mediated, making it a natural target for language model agents equipped with web search, case-law databases, and document parsing. If reliable, such agents could substantially reduce research time and help attorneys navigate large bodies of law. The central question is therefore not whether they can retrieve and reason, but whether they can do so reliably enough to be depended upon in practice.

However, most existing legal benchmarks report average correctness or partial credit, while legal research requires a standard closer to what lawyers need in practice: a legally correct answer supported by real, valid citations \citep{guha2023legalbench,pipitone2024legalbenchrag,fei2024lawbench,li2025legalagentbench}. To address this gap, we introduce \textbf{Legal Research Bench} (LRB), which counts a response as correct only if it gives the required legal answer and its cited authorities verify.

\begin{figure}[t]
\centering
\includegraphics[width=0.95\linewidth]{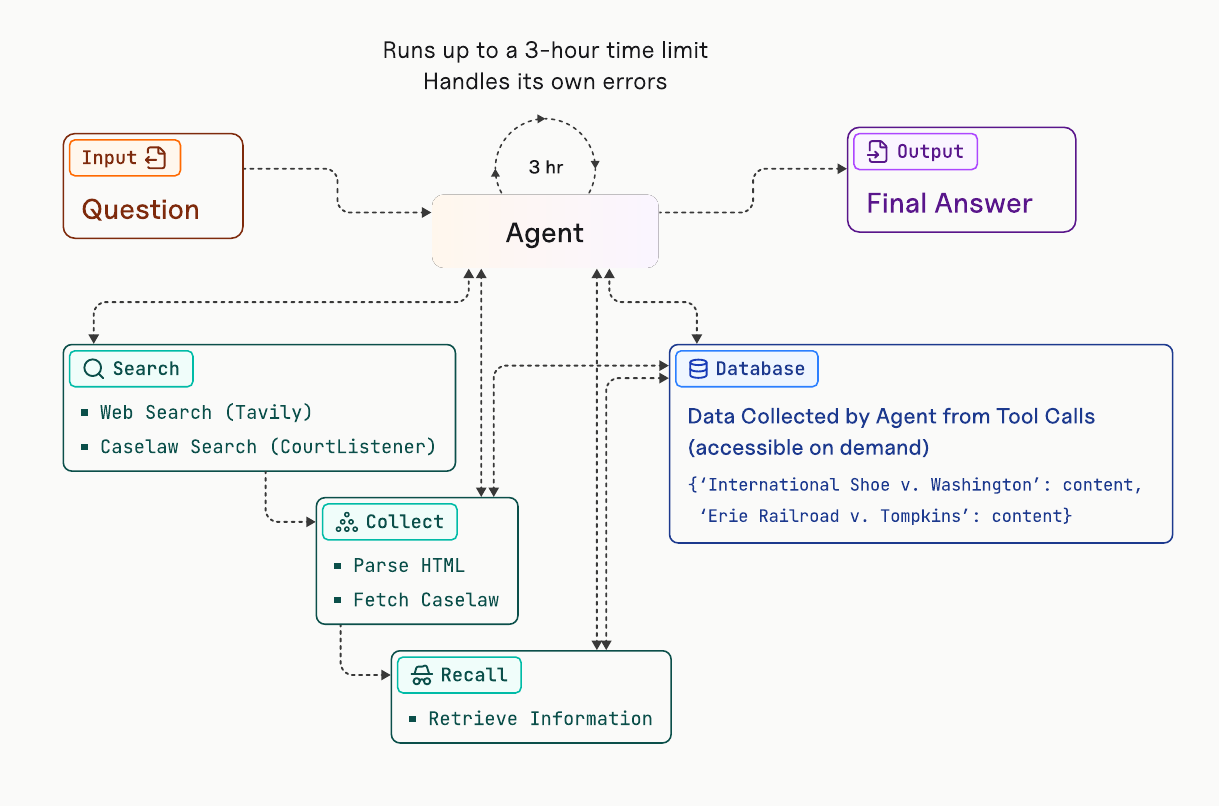}
\caption{Overview of the LRB evaluation setup. Each question is handed to a
tool-using agent that can search the web (Tavily) and case law (CourtListener),
parse and fetch pages, and recall content it has collected into a working
database, before submitting a final answer. Runs are capped at three hours, and
the agent manages its own errors.}
\label{fig:overview}
\end{figure}

Our main contributions are the following:
\begin{itemize}
    \item \textbf{Benchmark.} We introduce LRB, a 413-question benchmark authored by experts for open-ended U.S. legal research tasks, with gold answers, verified authorities, and binary rubrics.
    \item \textbf{Harness.} We release a reproducible agentic evaluation harness in which models search the web and case law, parse and fetch pages, recall retrieved content, and submit final answers. Figure~\ref{fig:overview} gives an overview of the evaluation setup.
    \item \textbf{Metric.} We evaluate reliability with all-pass grading: a response is correct only if it satisfies every rubric criterion and every cited authority verifies.
    \item \textbf{Judge validation.} We validate an LLM judge against expert attorneys before using it to score model outputs.
    \item \textbf{Empirical results.} We evaluate thirteen frontier models using this harness and find high partial credit but limited all-pass reliability: the best model attains 42.9\% all-pass, failures vary systematically across areas of law, reconciliation questions are especially difficult, and additional agentic effort does not predict higher accuracy.
\end{itemize}

% =====================================================================

\section{Related work}

\paragraph{Legal benchmarks.} Most legal NLP benchmarks target static, single-turn tasks: LegalBench \citep{guha2023legalbench}, LegalBench-RAG \citep{pipitone2024legalbenchrag}, LawBench \citep{fei2024lawbench}, and PRBench \citep{akyurek2025prbench} test legal reasoning, retrieval, and domain knowledge, but generally do not measure whether a tool-using agent can produce a complete final research answer over primary authority. Recent agent benchmarks move closer: LegalAgentBench \citep{li2025legalagentbench} studies tool use in the Chinese legal domain, and Harvey's Legal Agent Benchmark \citep{grupen2026lab}, concurrent with our work, evaluates long-horizon U.S. client-matter tasks under all-pass scoring. Its independent adoption of all-pass grading supports our metric choice, but the task setting differs: Harvey uses bounded matter-specific document sets, whereas LRB evaluates open-ended legal research across U.S. legal authority that an agent must locate, validate, and synthesize. Prior work on legal hallucination further motivates authority-grounded scoring: commercial legal AI tools still hallucinate on 17 to 33 percent of queries \citep{magesh2024hallucination}, and general-purpose models fabricate authority far more often \citep{dahl2024large}.

\paragraph{Long-horizon agents.} Benchmarks such as GAIA \citep{mialon2024gaia}, SWE-bench \citep{jimenez2024swe}, WebArena \citep{zhou2024webarena}, and $\tau$-bench \citep{yao2024taubench} study multi-step, tool-grounded behavior rather than single-turn accuracy. These settings make trajectory matching a poor target: different paths can reach the same answer, and similar trajectories can produce different outcomes. LRB follows this outcome-level view, scoring the submitted answer rather than the
path taken to it.

\paragraph{Judges and measurement validity.} Grading open-ended answers at scale increasingly relies on model-mediated judging. The LLM-as-judge paradigm \citep{zheng2023judging} shows that strong models can approximate human judgments, and related work has used model judges for criteria-based open-ended assessment and rubric-based grading \citep{liu2023g,hashemi2024llm}. But judge scores are not self-validating: they must agree with human or expert labels for the construct being measured, and their errors must not distort the benchmark's primary metric. Measurement-focused work likewise emphasizes explicit constructs and measurement procedures \citep{liang2022helm,bowman2021will,jacobs2021measurement}. LRB treats legal research reliability as a construct to be measured directly using all-pass grading rather than partial credit, and validates its judge against expert attorneys before scoring.

% =====================================================================
\section{Legal Research Bench}

\subsection{Data collection}
Each LRB item is authored by practicing U.S. attorneys and consists of an
open-ended legal research question, a gold answer, a curated source list, and a
binary grading rubric (Section~\ref{sec:metrics}). Items are designed to require
legal reasoning rather than factual lookup, are written around novel fact patterns
to reduce memorization, and are included only when the correct answer is objective
enough for attorney agreement. When legal authority may change over time, questions
are temporally anchored with an explicit answer date (e.g.\ ``Answer as of
[date]''), and the agent is prompted to answer as of that date. Each item is
independently reviewed by at least one additional attorney, and ground-truth
materials are written without AI assistance beyond light copyediting. In total, 15
practicing attorneys, with an average of 15 years of legal experience, contributed
to authoring and review, and building each item required approximately four
attorney-hours.

\subsection{Dataset}
\label{sec:dataset}
LRB comprises 413 questions spanning eight areas of law and ten question types.
The corpus is partitioned into a 5-question public sample, a 200-question
validation set, and a 208-question test set. We publicly release the five-question
sample together with the full agent harness\footnote{\url{https://github.com/vals-ai/legal-research-bench}}
to support reproducibility. The validation and test questions are held out to
limit contamination. Unless noted otherwise, results in this paper are reported
over the full 413-question corpus.

Questions are multi-label: 37.8\% carry more than one area of law (1.41 areas on average). For compact reporting, we group the ten fine-grained question types into three overlapping \textbf{primary reasoning categories} and two \textbf{difficulty attributes}. The primary reasoning categories describe the main body of authority a question turns on: Statutory Interpretation, Doctrinal Rule Reasoning, or Regulatory Framework Interpretation. The difficulty attributes mark questions that are harder for a reason independent of the primary category: Reconciliation, where a question requires synthesizing multiple or conflicting authorities, and Temporal Validity, where the answer depends on whether a rule is still good law. Because these labels overlap, they are descriptive marginals rather than a partition; Appendix~\ref{app:qtypes} lists the ten underlying question types and their mapping to this scheme. Table~\ref{tab:dataset} gives the full composition.

Each question carries a curated source list (6.2 sources on average, median 6,
up to 23) and a binary rubric (9.4 items on average, median 9, up to 31). The
gold-standard answers average 536 words (median 429) in length.

\begin{table}[t]
\centering
\small
\begin{minipage}[t]{0.50\linewidth}
\centering
\begin{tabular}{lr}
\toprule
\textbf{Area of Law} & \textbf{N} \\
\midrule
Administrative / Regulatory   & 128 \\
Criminal                      & 117 \\
Business \& Commercial        &  88 \\
Constitutional / Civil Rights &  72 \\
Health                        &  58 \\
Civil Litigation              &  51 \\
Family                        &  45 \\
Immigration                   &  22 \\
\bottomrule
\end{tabular}
\end{minipage}\hfill
\begin{minipage}[t]{0.48\linewidth}
\centering
\begin{tabular}{lr}
\toprule
\textbf{Question Type} & \textbf{N} \\
\midrule
\multicolumn{2}{l}{\emph{Primary reasoning categories}} \\
Statutory Interpretation        & 288 \\
Doctrinal Rule Reasoning        & 262 \\
Regulatory Framework Interp.    &  91 \\
\midrule
\multicolumn{2}{l}{\emph{Difficulty attributes}} \\
Reconciliation                  & 116 \\
Temporal Validity               &  66 \\
\bottomrule
\end{tabular}
\end{minipage}
\caption{Dataset composition over all 413 questions: distribution across the
eight areas of law (left) and the question-type scheme of three overlapping
primary reasoning categories and two difficulty attributes (right). Questions are
multi-label, so the counts sum to more than 413.}
\label{tab:dataset}
\end{table}

\subsection{Rubric and metrics}
\label{sec:metrics}
Each question is graded against binary rubric criteria that the authoring attorneys judged
individually necessary and jointly sufficient for a correct answer. Satisfying every
criterion is therefore equivalent to producing a correct answer by that expert standard. Criteria are
weighted [+1], [+2], or [+3], with at least one [+3] item capturing the central
legal conclusion. We report \textbf{all-pass} as the primary metric: a response is
correct only if \emph{every} rubric criterion is satisfied. \textbf{Weighted
pass}, the share of available rubric weight a response earns, is reported only as
a secondary, partial-credit measure. Grounding is part of correctness: the rubric
specifies which authorities a correct answer must cite, and any invalid cited URL
fails a \textbf{source check} that marks the entire response wrong. Following recent recommendations to report statistical uncertainty in language model evaluations \citep{miller2024adding}, we report standard errors for leaderboard estimates and confidence intervals clustered by question for category-level analyses; Appendix~\ref{app:qtypes} details the procedure.

% =====================================================================
\section{Experimental setup}

\subsection{Agents}
Models are evaluated as tool-using agents (Figure~\ref{fig:overview}). Each agent has access to five tools. Web search via Tavily \citep{tavily} returns links with
short excerpts. Case-law search via CourtListener \citep{courtlistener} finds court
opinions by keyword, by natural language, or by citation lookup, and fetches their full
text; page parsing does the same for any other web page. Both store what they fetch under
a key the agent chooses, and retrieval then answers a prompt against the stored documents,
so an opinion too long to hold in context can still be used. A final tool submits the
answer. Runs are capped by a three-hour wall-clock budget per question rather than by a fixed number of turns. A time budget better reflects real research workflows and makes the evaluation comparable across alternative harnesses, tools, and agent implementations. The cap is far from binding in practice: no fully correct (all-pass) answer took longer than 1.9 hours, and 95\% of all runs complete within roughly 50 minutes. We evaluate $N=13$ frontier models on the corpus, all released between February and June 2026.

\subsection{Judge calibration}
\label{sec:judge}

Since every score in LRB is assigned by a single LLM judge, we validate the judge against expert attorneys before using it for model evaluation. In a held-out calibration study, three attorneys and three candidate judges (Claude Sonnet 4.6, Gemini 3.1 Pro Preview, and GPT-5.4) scored 341 rubric items drawn from 45 model responses to 15 questions. The attorney majority serves as the reference label, and human inter-rater agreement sets the baseline: attorneys agree with the majority label on 83.4\% of rubric items, with Cohen's $\kappa = 0.644$. All three candidate judges meet or exceed this baseline, with $\kappa$ ranging from 0.708 to 0.726.

We select GPT-5.4 as the benchmark judge for two reasons. First, its pass rate is closest to the attorney majority, reducing the risk that benchmark scores are systematically shifted by a judge that is much stricter or more lenient than expert graders. Second, its false positives and false negatives are more balanced than those of the other candidate judges, at 5.6\% and 7.0\%, respectively. This balance matters under all-pass grading, where one-sided item-level errors can compound across multi-item rubrics. Aggregation at the task level does not degrade this agreement: translating GPT-5.4's per-item verdicts into the all-pass score, it agrees with the expert majority on 86.7\% of responses ($\kappa = 0.710$), close to the human inter-rater baseline of 87.9\% ($\kappa = 0.735$). Table~\ref{tab:judge} reports the calibration results, and Appendix~\ref{app:judge} gives additional robustness checks.

\begin{table}[t]
\centering
\begin{tabular}{lrrrrr}
\toprule
\textbf{Rater} & \textbf{Agree} & \textbf{Cohen's $\kappa$} & \textbf{FP} &
\textbf{FN} & \textbf{Pass} \\
\midrule
\textbf{GPT-5.4} (selected)   & 87.4 & 0.715 & 5.6 &  7.0 & 66.3 \\
Gemini 3.1 Pro Preview        & 87.4 & 0.726 & 2.9 &  9.7 & 61.0 \\
Claude Sonnet 4.6             & 86.5 & 0.708 & 3.2 & 10.3 & 60.7 \\
\midrule
Human (inter-rater avg)       & 83.4 & 0.644 & n/a & n/a  & 67.7 \\
\bottomrule
\end{tabular}
\caption{Judge--expert agreement vs.\ the human inter-rater baseline, over 341
rubric-item comparisons. Agree and $\kappa$ score each judge against the attorney majority; FP and FN are the false positive and false negative rates, respectively; Pass is the share of items a rater marks pass. The Human row is
pairwise attorney-to-attorney, so FP and FN are undefined, and its Pass entry is the majority's. All three judges exceed the human agreement baseline; GPT-5.4 is selected for its balanced error profile and high agreement with the human expert majority.}
\label{tab:judge}
\end{table}

% =====================================================================
\section{Results}

\subsection{Reliability remains limited}
Table~\ref{tab:leaderboard} reports overall performance across the full corpus. Claude Opus 4.8 achieves the highest all-pass rate at 42.9\%, followed by GPT-5.5 at 40.0\% and Claude Sonnet 4.6 at 38.5\%; the weakest model reaches 12.6\%. Weighted pass is much higher, from 68\% to 86\%: agents recover most of an answer while still missing at least one element it requires. The distance, between nearly correct and complete, can be crucial in a legal setting.

\begin{table}[t]
\centering
\begin{tabular}{lrrrrrr}
\toprule
\textbf{Model} & \textbf{All-pass} & \textbf{Wtd pass} & \textbf{Sources} &
\textbf{Turns} & \textbf{Tools} & \textbf{\$/test} \\
\midrule
\textbf{Claude Opus 4.8}   & $\mathbf{42.9{\pm}2.4}$ & $85.4{\pm}1.1$ &  8.6 & 11.7 &  28.7 & 2.97 \\
GPT-5.5                    & $40.0{\pm}2.4$ & $83.4{\pm}1.1$ &  9.0 & 58.5 &  66.8 & 7.34 \\
Claude Sonnet 4.6          & $38.5{\pm}2.4$ & $85.6{\pm}1.0$ & 10.8 & 25.2 &  52.1 & 2.46 \\
GLM-5.2                    & $31.2{\pm}2.3$ & $82.7{\pm}1.1$ &  8.5 & 23.3 &  45.1 & 0.91 \\
Gemini 3.5 Flash           & $30.8{\pm}2.3$ & $83.2{\pm}1.0$ &  7.9 & 47.0 &  47.0 & 1.37 \\
MiniMax-M3                 & $29.8{\pm}2.3$ & $80.1{\pm}1.2$ &  8.9 & 26.6 &  52.6 & 0.36 \\
GLM-5.1                    & $27.6{\pm}2.2$ & $76.8{\pm}1.2$ &  7.6 & 25.3 &  43.9 & 0.39 \\
Qwen3.7-Max                & $25.2{\pm}2.1$ & $78.4{\pm}1.2$ &  7.0 & 19.7 &  37.9 & 1.44 \\
DeepSeek-V4-Pro            & $23.2{\pm}2.1$ & $80.9{\pm}1.1$ &  6.2 & 32.8 &  50.7 & 0.74 \\
Gemini 3.1 Pro Preview     & $20.8{\pm}2.0$ & $75.4{\pm}1.2$ &  4.0 & 20.7 &  21.5 & 0.79 \\
Kimi K2.6                  & $15.5{\pm}1.8$ & $75.3{\pm}1.2$ &  4.7 & 99.6 & 112.3 & 1.74 \\
Grok 4.3                   & $15.3{\pm}1.8$ & $68.8{\pm}1.3$ &  4.0 & 14.1 &  24.4 & 0.24 \\
GPT-5.4-mini               & $12.6{\pm}1.6$ & $68.0{\pm}1.3$ &  4.4 & 46.6 &  94.4 & 1.90 \\
\bottomrule
\end{tabular}
\caption{Leaderboard over the full 413-question corpus, sorted by all-pass. Pass
columns are mean $\pm$ standard error (\%); Sources is the mean number of authorities the model cites ; Turns and Tools are mean agent turns and tool calls per question;
\$/test is mean cost per question in USD. Appendix~\ref{app:paired} compares
adjacent entries on paired per-question outcomes.}
\label{tab:leaderboard}
\end{table}

\subsection{Partial correctness hides substantive omissions}
\label{sec:failures}
All-pass failures are not confined to minor omissions. Across the corpus, models miss at least one high-value [+3] criterion on 29.5\% of questions, and these misses account for roughly 41\% of all failures (Appendix~\ref{app:additional}, Figure~\ref{fig:failures}). The remaining failures are also often substantive: lower-weight criteria require a specific statute, rule, controlling authority, or legal interpretation. A further 4.3\% of runs satisfy every rubric criterion but fail the source check, because a cited URL is invalid or does not point to the authority claimed. Thus, the gap between weighted pass and all-pass reflects missing required elements of the legal research answer, not a long tail of minor details.

\subsection{Reliability varies systematically}
\label{sec:results-domains}
\begin{figure}[t]
\centering
\includegraphics[width=\linewidth]{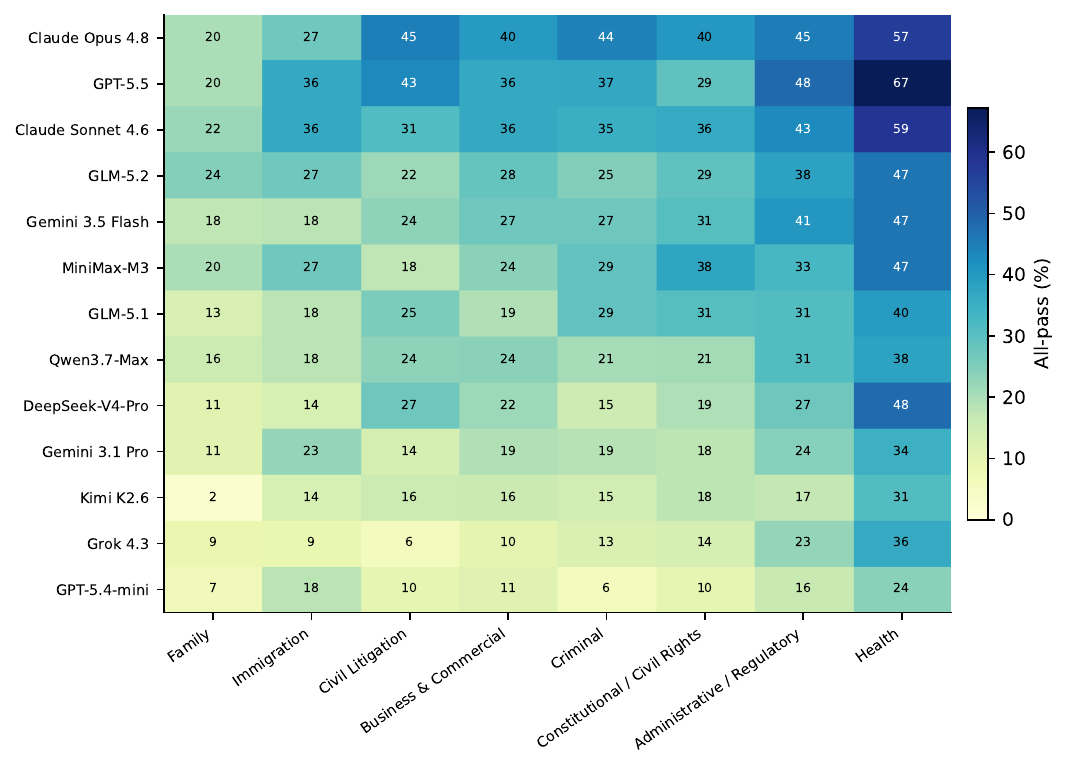}
\caption{All-pass rate by area of law.}
\label{fig:area-heatmap}
\end{figure}

\begin{figure}[t]
\centering
\includegraphics[width=0.58\linewidth]{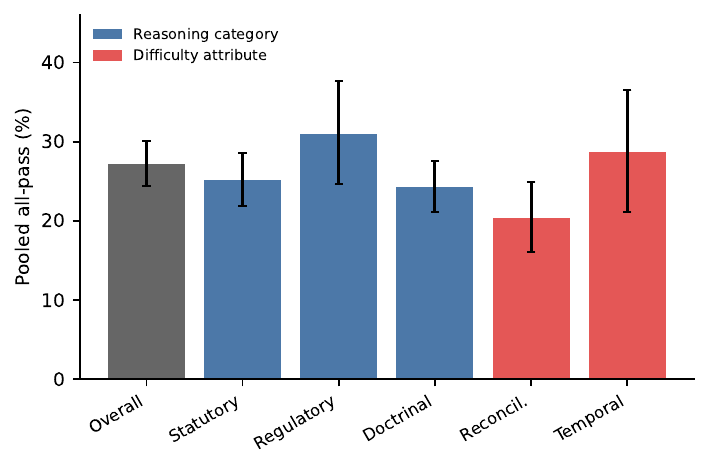}
\caption{Pooled all-pass rate by primary reasoning category (blue) and difficulty
attribute (red), with 95\% confidence intervals. The reasoning categories cluster
near the overall rate, whereas reconciliation is markedly harder.}
\label{fig:type-pooled}
\end{figure}

Reliability varies by practice area. Averaged across models, all-pass
ranges from 44.2\% in Health and 32.2\% in Administrative/Regulatory to 14.9\% in
Family, with most other areas in the low-to-mid 20s
(Figure~\ref{fig:area-heatmap}). No model is strong everywhere: the best model in
the hardest area reaches only 24.4\% all-pass. Difficulty also varies by question
type (Figure~\ref{fig:type-pooled}). Reconciliation questions, which require
synthesizing multiple or conflicting authorities, pass at 20.4\% (95\% CI [16.1, 24.9]) 
versus 29.9\% ([26.4, 33.4]) for questions without the attribute, a gap of 9.4 percentage points, a penalty that holds for every model, while
temporal-validity questions sit near the benchmark average at 28.7\%. The
reconciliation penalty is not an artifact of rubric length: reconciliation
questions carry slightly fewer criteria than the rest, and the effect persists in a
logistic model controlling for rubric length and source count. Appendix~\ref{app:qtypes}
gives the per-model breakdown and the regression with these controls.

\subsection{More turns, tool calls, and cost do not predict reliability}
\label{sec:effort}
Higher turn counts, inference cost, and tool-call counts do not reliably predict correctness across models (Figure~\ref{fig:effort}; tool calls in Appendix~\ref{app:additional}, Figure~\ref{fig:effort-tools}). Claude Opus 4.8 achieves the highest all-pass rate while using the fewest turns of any model (11.7), whereas Kimi K2.6 averages nearly 100 turns and 112 tool calls but reaches only 15.5\% all-pass. Cost shows the same pattern: GPT-5.5 is the most expensive run (\$7.34 per question) yet trails Claude Opus 4.8, which scores higher at roughly 40\% of the cost. Per-question trajectories (Appendix~\ref{app:traj}) instead show stronger agents reading and verifying authority rather than repeatedly issuing searches. Appendix~\ref{app:verbosity} reports answer length, the one effort measure
that does correlate with all-pass.

\begin{figure}[t]
\centering
\includegraphics[width=\linewidth]{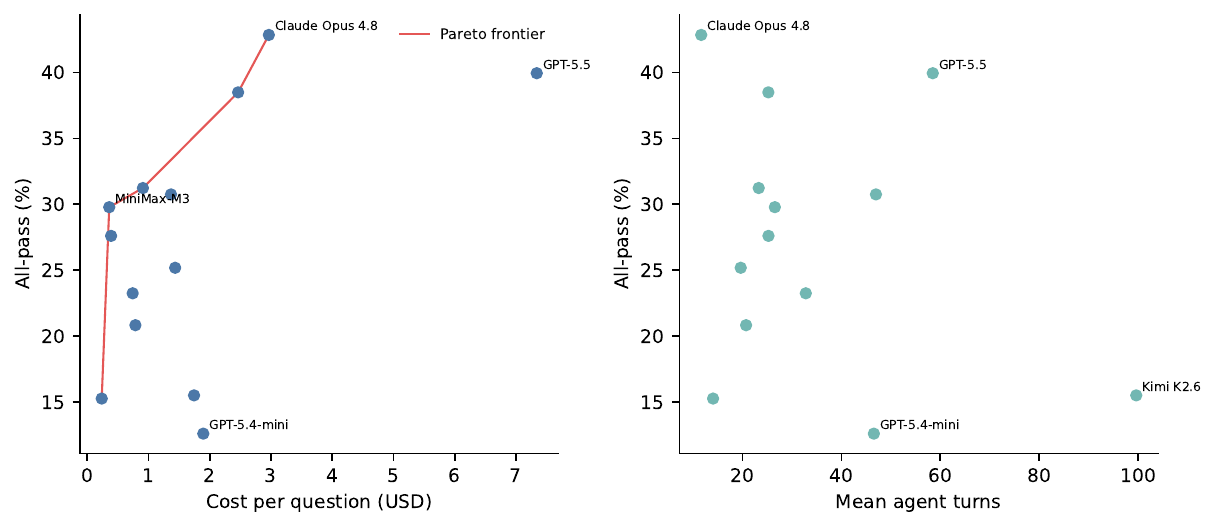}
\caption{All-pass against inference cost (left) and mean agent turns (right).
The red line on the cost panel marks the Pareto frontier: the models for which
no other model is both cheaper and more accurate. Tool calls are plotted separately in
Appendix~\ref{app:additional}, Figure~\ref{fig:effort-tools}.}
\label{fig:effort}
\end{figure}

\subsection{Tool access is necessary but not sufficient}
\label{sec:ablation}
In a no-tool, single-shot setting,
performance collapses across all thirteen models (Table~\ref{tab:ablation}):
all-pass falls by 22.5 points on average, and every tool-using model outperforms
every no-tool model. The strongest models lose the most (Claude Opus 4.8 falls
34.1 points), indicating that their advantage comes from locating and reading
authority rather than from memorized law. Combined with Section~\ref{sec:effort},
this indicates that legal research ability rests on tool-grounded retrieval and
verification, yet once tools are available, using them moredoes not predict higher reliability.

\begin{table}[t]
\centering
\begin{tabular}{lrrrr}
\toprule
\textbf{Model} & \textbf{All-pass (tools)} & \textbf{All-pass (1-shot)} &
\textbf{Wtd (tools)} & \textbf{Wtd (1-shot)} \\
\midrule
Claude Opus 4.8        & 42.9 &  8.7 & 85.4 & 63.0 \\
GPT-5.5                & 40.0 &  6.8 & 83.4 & 57.9 \\
Claude Sonnet 4.6      & 38.5 &  7.3 & 85.6 & 58.0 \\
GLM-5.2                & 31.2 &  3.4 & 82.7 & 50.3 \\
Gemini 3.5 Flash       & 30.8 &  6.3 & 83.2 & 60.8 \\
MiniMax-M3             & 29.8 &  2.9 & 80.1 & 48.4 \\
GLM-5.1                & 27.6 &  3.1 & 76.8 & 46.7 \\
Qwen3.7-Max            & 25.2 &  2.7 & 78.4 & 53.0 \\
DeepSeek-V4-Pro        & 23.2 &  4.1 & 80.9 & 52.0 \\
Gemini 3.1 Pro Preview & 20.8 &  6.5 & 75.4 & 60.8 \\
Kimi K2.6              & 15.5 &  2.9 & 75.3 & 48.4 \\
Grok 4.3               & 15.3 &  4.1 & 68.8 & 49.6 \\
GPT-5.4-mini           & 12.6 &  2.2 & 68.0 & 42.6 \\
\bottomrule
\end{tabular}
\caption{Ablation over the full 413-question corpus: each model run with its full
tool environment versus single-shot with no tool access, sorted by all-pass with
tools. All values are percentages. Removing tools lowers all-pass by 22.5 points
and weighted pass by 25.6 points on average, and every tool-using model outscores
every single-shot one.}
\label{tab:ablation}
\end{table}

% =====================================================================

\section{Limitations and future work}

LRB is limited to U.S., English-language legal research and to a fixed snapshot of models, tools, and legal authority.\footnote{A live leaderboard is maintained at \url{https://www.vals.ai/benchmarks/legal_research}.} Its questions are also designed to have objective answers, which is necessary for all-pass grading but excludes parts of legal work that involve strategy, judgment under uncertainty, client counseling, negotiation, or drafting choices with no single correct answer. Although we validate the LLM judge against expert attorneys before using it for scoring, the reported results still depend on a model-mediated grading procedure rather than full expert adjudication of every response. Finally, each model was evaluated in a single run per question, so the reported rates carry sampling uncertainty over questions but not run-to-run variance. Future work should extend this evaluation design to additional jurisdictions, languages, and legal tasks; study how reliability changes under different tool environments and human-in-the-loop workflows; and develop evaluation protocols that measure not only final-answer correctness, but also when agents appropriately surface uncertainty, ask for clarification, or defer to human review.

\section{Conclusion}

LRB makes end-to-end legal research reliability measurable by pairing all-pass grading of final answers with judge validation against expert attorneys. Under this evaluation, the frontier tool-using models we test often earn substantial partial credit, but the strongest model at the time of evaluation reaches only 42.9\% all-pass. The observed failures are systematic rather than incidental: all-pass rates vary across practice areas, and reconciliation tasks are especially difficult. Across models, higher turn counts and tool-call counts do not reliably predict higher accuracy. LRB therefore suggests that legal agents should not be judged by whether they can produce plausible research, but by whether they can consistently produce work that lawyers could safely rely on.

\section*{Acknowledgments}
We thank the practicing attorneys who authored and reviewed the LRB questions, gold
answers, and rubrics, and took part in the judge calibration study.

\section*{Ethics Statement}
Legal research is high-stakes and highly sensitive work: errors can affect
liberty, livelihood, and legal rights, and a single fabricated authority or
misstated holding can cause real harm. Our central finding is that current agents
are not yet reliable at this task, and we present LRB as a measurement of that gap
rather than as an endorsement of any system for practice. The benchmark is not a
certification of fitness for deployment, and its results should not be read as
clearing models for unsupervised legal use; agent outputs are not legal advice and
should be reviewed by a licensed attorney before they are relied upon. We caution
in particular against automation bias, since the fluent, well-structured answers
these agents produce can appear authoritative even when the central legal
conclusion is wrong.

The benchmark itself raises no human-subjects or privacy concerns. All questions
are original hypotheticals authored by the participating attorneys around novel
fact patterns; they contain no real client information, no personally identifiable
information, and no privileged or confidential material. The attorneys who wrote
and reviewed the items did so as compensated contributors and consented to the
use of their work. We release a small public sample and the evaluation harness to
support scrutiny and reproducibility, while holding out the remaining questions to
limit contamination.

\bibliography{colm2026_conference}
\bibliographystyle{colm2026_conference}

\appendix
\section{Sample question}
\label{app:sample}
Below is a sample question from the public dataset 
split, P-005 (Business \& Commercial; Statutory Interpretation), used as the
running example in Figure~\ref{fig:trajectory}. This question is graded against a binary rubric of 12 weighted criteria.

\paragraph{Question.}
Client is a Virginia corporation (``Client'') that hauls refrigerated freight out of Rockbridge County.  In early 2024, Client needed two new reefer trailers and went to a national manufacturer (``Manufacturer'') to pick out a model and negotiate the refrigeration specs.  Client then arranged for a Virginia equipment financing company (``Lessor'') to buy the trailers (``Trailers'') from Manufacturer for \$85,000 each and lease them to Client under a written 60-month lease (``Lease'') at \$2,200 per month per trailer, and Lessor gave Client a complete copy of its purchase agreement with Manufacturer, which included Manufacturer's standard two-year warranty on all mechanical and refrigeration components, before Client signed.

Four months after delivery, both Trailers had compressor failures that left the units unable to hold temperature, and Manufacturer agreed to fix them under the warranty but filed for Chapter 7 two months later, where the trustee confirmed it wouldn't honor outstanding warranty claims.  That said, the Lease doesn't include any warranties from Lessor and tells Client to take equipment claims to Manufacturer.  It also says that Client's payments are ``absolute and unconditional, irrespective of any defense or any right of setoff, counterclaim, or recoupment.''  Client has stopped paying Lessor.  Given these facts, can Client stop its payments to Lessor, and is Lessor responsible for the defective Trailers?

\section{Judge calibration: additional analyses}
\label{app:judge}
\paragraph{GPT-5.4 versus the other judge candidates}
Gemini 3.1 Pro Preview attains the top agreement ($\kappa = 0.726$), marginally
above GPT-5.4 ($0.715$), but we do not select on $\kappa$ alone. GPT-5.4 has the most balanced error profile, with a false-positive to false-negative ratio
of about $1{:}1.3$, versus roughly $1{:}3.2$ to $1{:}3.3$ for the stricter Claude
and Gemini judges (Table~\ref{tab:judge}); a balanced profile matters under
all-pass grading, where one-sided item-level errors compound across a multi-item
rubric. Its pass rate (66.3\%) is also closest to the expert majority (67.7\%). Because the selected judge remains mildly strict (false
negatives modestly exceed false positives), reported all-pass rates are a slightly
conservative estimate of human-graded performance.

Beyond the headline agreement in Section~\ref{sec:judge}, the alignment study
supports the selected judge in two additional ways. First, the agreement is not an
artifact of easy questions: excluding four near-uniform questions whose constant
labels make $\kappa$ undefined leaves 308 comparisons on the harder items, and
every judge's $\kappa$ falls by less than one point (GPT-5.4: $0.715 \to 0.705$).
Second, there is no same-family leniency: each judge's \emph{lowest} agreement is
on models from its own provider.

\section{\texorpdfstring{Fine-grained question types and adjusted category effects}{Fine-grained question types and adjusted category effects}}
\label{app:qtypes}
Every question is answered by all thirteen models, so observations are clustered
within questions. Pooled category-level and area-level rates use 95\% confidence
intervals from a question-level cluster bootstrap, resampling questions with replacement
and carrying all thirteen of a resampled question's model outcomes with it. The
intraclass correlation of all-pass within question is 0.40, a design effect of 5.8 at
thirteen models. Per-model standard errors in Table~\ref{tab:leaderboard} are computed
directly, since each model contributes one observation per question.

In our analysis, ten fine-grained question types are grouped into three
overlapping primary reasoning categories (the primary source of authority a
question turns on) and two difficulty attributes (orthogonal sources of
difficulty). Table~\ref{tab:qtypes} lists the ten types, the category or attribute
each maps to, and the pooled all-pass
rate over the full 413-question corpus. Questions are multi-label, so counts sum
to more than 413 and the groups are descriptive marginals rather than a
partition.

\begin{table}[h]
\centering
\begin{tabular}{llrrc}
\toprule
\textbf{Question type} & \textbf{Maps to} & \textbf{N} & \textbf{All-pass} & \textbf{95\% CI} \\
\midrule
\multicolumn{5}{l}{\emph{Primary reasoning categories}} \\
Statutory Interpretation                   & Statutory  & 288 & 25.2 & [21.9, 28.5] \\
Application of Precedent to a Fact Pattern  & Doctrinal  & 240 & 23.3 & [20.1, 26.6] \\
Doctrinal Test Identification               & Doctrinal  & 116 & 24.3 & [19.5, 29.3] \\
Regulatory Framework Interpretation         & Regulatory &  91 & 31.0 & [24.6, 37.6] \\
\midrule
\multicolumn{5}{l}{\emph{Difficulty attributes}} \\
Cross-Jurisdiction Comparison               & Reconciliation     & 48 & 21.5 & [14.6, 28.8] \\
Consensus vs.\ Disagreement Across Courts   & Reconciliation     & 27 & 17.9 & [8.5, 28.2] \\
Resolving Conflicting Precedent             & Reconciliation     & 25 & 25.6 & [14.2, 37.8] \\
Interaction of Multiple Legal Regimes       & Reconciliation     & 71 & 21.4 & [16.0, 27.2] \\
Validity of Precedent                       & Temporal Validity  & 42 & 29.0 & [20.0, 38.7] \\
Timeline / Doctrinal Evolution              & Temporal Validity  & 45 & 27.2 & [18.3, 36.5] \\
\bottomrule
\end{tabular}
\caption{The ten question types, their mapping to the category/attribute scheme used in
Section~\ref{sec:results-domains}, and pooled all-pass over the full corpus
(percentages, with question-level
cluster-bootstrap 95\% confidence intervals, which account for every question being
answered by all thirteen models). N is the number of
questions carrying each type. }
\label{tab:qtypes}
\end{table}

\begin{figure}[h]
\centering
\includegraphics[width=0.7\linewidth]{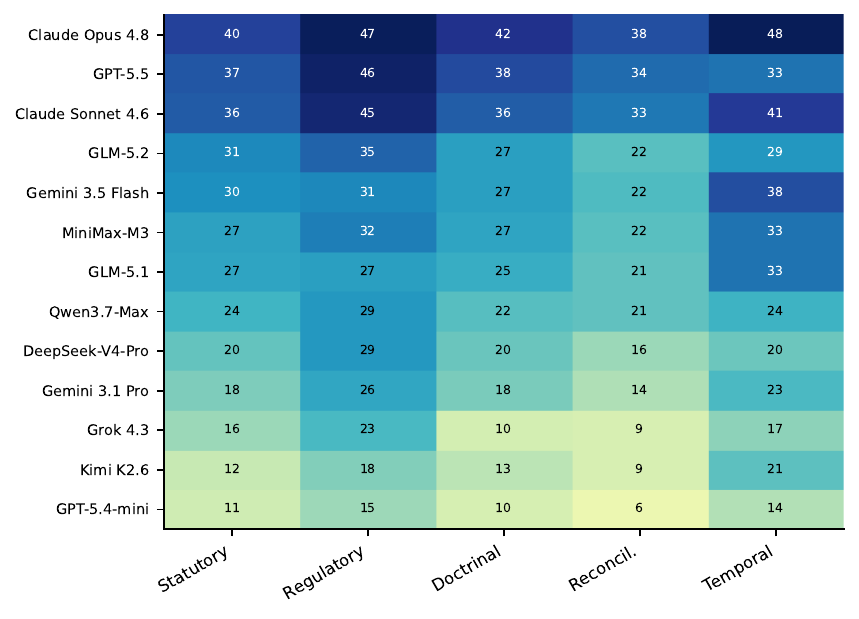}
\caption{Per-model all-pass rate by primary reasoning category and difficulty
attribute.}
\label{fig:type-heatmap}
\end{figure}

Figure~\ref{fig:type-heatmap} breaks the pooled rates of Section~\ref{sec:results-domains}
down by model, across the three primary reasoning categories and the two difficulty
attributes. The categories differ little across models, all sitting near the 27.2\%
overall rate, while the Reconciliation attribute is consistently the clearest
difficulty signal.

Rubric length varies across questions and is itself a predictor of all-pass, so
here we check whether the reconciliation penalty reported in
Section~\ref{sec:results-domains} is driven by it. Pooled across all thirteen models,
all-pass falls from 41.8\% on questions carrying six or fewer criteria to 15.2\% on
questions carrying thirteen or more, a rank correlation of $-0.287$ between a
question's criterion count and its all-pass rate. To test whether this accounts for the
penalty, we regress all-pass on a reconciliation indicator, rubric length, and source
count, with area-of-law indicators as further controls. All eight areas enter
without an omitted reference, since questions can carry more than one area and the
indicators therefore do not sum to a constant.
Table~\ref{tab:adjusted} compares the unadjusted and adjusted odds ratios.

\begin{table}[h]
\centering
\begin{tabular}{lrr}
\toprule
\textbf{Predictor} & \textbf{Unadjusted OR} & \textbf{Adjusted OR} \\
\midrule
Reconciliation                    & 0.60 & 0.68 \\
\midrule
Rubric length (per criterion)     & ---  & 0.89 \\
Source count (per source)         & ---  & 0.93 \\
\bottomrule
\end{tabular}
\caption{All-pass odds ratios for reconciliation before and after adjusting for
rubric length and source count. Area-of-law indicators are included in the adjusted
model as controls but are not reported. Rubric length and source count enter only the
adjusted model.}
\label{tab:adjusted}
\end{table}

An odds ratio of one would mean no difference. Adjusting for rubric length and
source count moves the reconciliation effect from 0.60 to 0.68, so the penalty weakens
but stays below one (coefficient $-0.380$, 95\% CI
$[-0.73, -0.04]$). Both controls predict all-pass in their own right, at 0.89 per
additional rubric criterion and 0.93 per additional required source, yet neither
accounts for the penalty: reconciliation questions carry 9.0 criteria on average
against 9.5 for the rest, so the confound runs against the observed penalty rather
than with it.

\section{\texorpdfstring{Paired model comparisons}{Paired model comparisons}}
\label{app:paired}
Every model answers every question, so adjacent leaderboard entries can be
compared on paired per-question outcomes rather than through independent standard
errors. Table~\ref{tab:paired} reports McNemar's test for each of the twelve adjacent
pairs in Table~\ref{tab:leaderboard}. The discordant counts give the number of
questions the higher-ranked model passes and the lower-ranked model fails, followed by
the reverse. Only two adjacent pairs separate at the 5\% level, so the leaderboard
supports tiers rather than a strict ordering.

\begin{table}[h]
\centering
\small
\begin{tabular}{lrr}
\toprule
\textbf{Adjacent pair (all-pass \%)} & \textbf{Discordant} & \textbf{McNemar $p$} \\
\midrule
Claude Opus 4.8 (42.9) vs.\ GPT-5.5 (40.0)                & 52/39 & 0.173 \\
GPT-5.5 (40.0) vs.\ Claude Sonnet 4.6 (38.5)              & 50/44 & 0.536 \\
Claude Sonnet 4.6 (38.5) vs.\ GLM-5.2 (31.2)              & 57/27 & \textbf{0.001} \\
GLM-5.2 (31.2) vs.\ Gemini 3.5 Flash (30.8)               & 53/51 & 0.845 \\
Gemini 3.5 Flash (30.8) vs.\ MiniMax-M3 (29.8)            & 58/54 & 0.705 \\
MiniMax-M3 (29.8) vs.\ GLM-5.1 (27.6)                     & 51/42 & 0.351 \\
GLM-5.1 (27.6) vs.\ Qwen3.7-Max (25.2)                    & 49/39 & 0.286 \\
Qwen3.7-Max (25.2) vs.\ DeepSeek-V4-Pro (23.2)            & 50/42 & 0.404 \\
DeepSeek-V4-Pro (23.2) vs.\ Gemini 3.1 Pro (20.8)         & 51/41 & 0.297 \\
Gemini 3.1 Pro (20.8) vs.\ Kimi K2.6 (15.5)               & 43/21 & \textbf{0.006} \\
Kimi K2.6 (15.5) vs.\ Grok 4.3 (15.3)                     & 33/32 & 0.901 \\
Grok 4.3 (15.3) vs.\ GPT-5.4-mini (12.6)                  & 31/20 & 0.123 \\
\bottomrule
\end{tabular}
\caption{Adjacent-pair comparisons in leaderboard order, over the full
413-question corpus. Discordant counts are higher-ranked-only passes / lower-ranked-only
passes. Bold $p$-values mark the two pairs that separate at the 5\% level.}
\label{tab:paired}
\end{table}

\section{Additional results}
\label{app:additional}

\begin{figure}[h]
\centering
\includegraphics[width=0.75\linewidth]{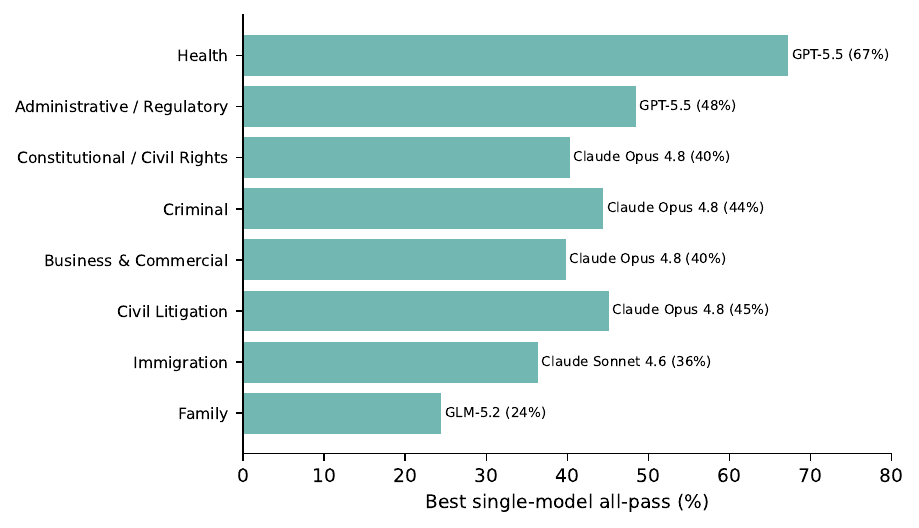}
\caption{Best per-model all-pass by area of law.}
\label{fig:area-leaders}
\end{figure}

\paragraph{Best model per area of law.}
Figure~\ref{fig:area-leaders} reports, for each area of law, the best single-model
all-pass rate, summarizing the heatmap of Figure~\ref{fig:area-heatmap} and
showing that no single model leads across all areas.

\begin{figure}[h]
\centering
\includegraphics[width=0.8\linewidth]{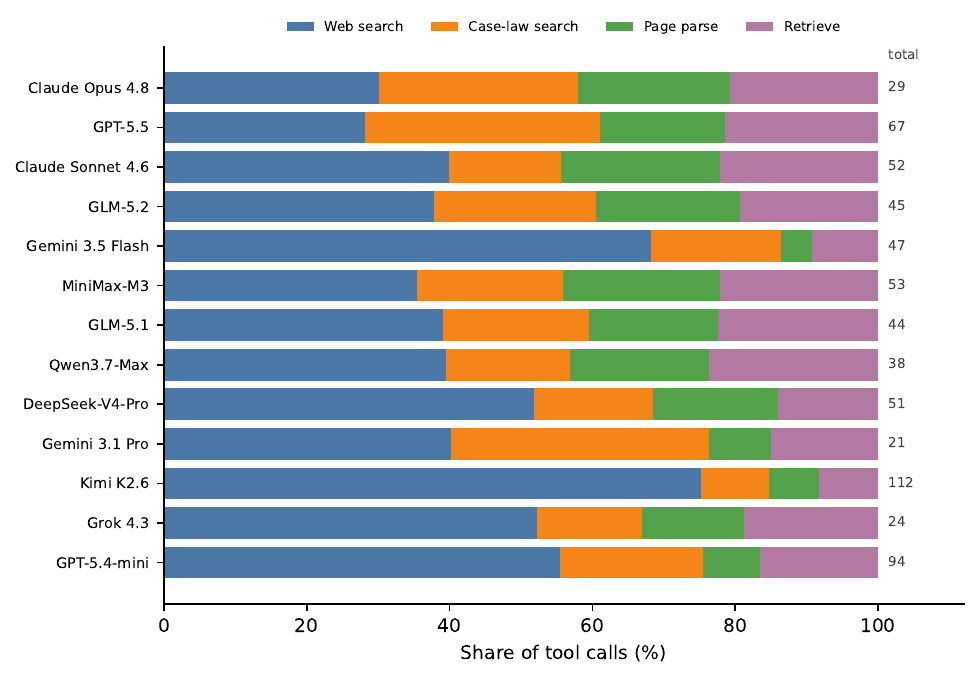}
\caption{Tool-usage composition by model. Totals at the right are mean tool calls
per question.}
\label{fig:tools}
\end{figure}

\paragraph{Tool-usage composition.}
Figure~\ref{fig:tools} breaks down each model's tool calls by type. Web search is
the most-used tool overall (about 27 calls per session against about 11 for
case-law search), but stronger models allocate more selectively: Claude Opus 4.8
balances case-law and web search in roughly equal proportion (about 8--9 calls
each). The heaviest searchers sit at the
bottom of the leaderboard, Kimi K2.6 at 86.4 web searches per question and GPT-5.4-mini
at 54.5, but call volume does not otherwise track accuracy: Gemini 3.1 Pro Preview issues
the fewest calls of any model and ranks tenth.

\begin{figure}[h]
\centering
\includegraphics[width=0.62\linewidth]{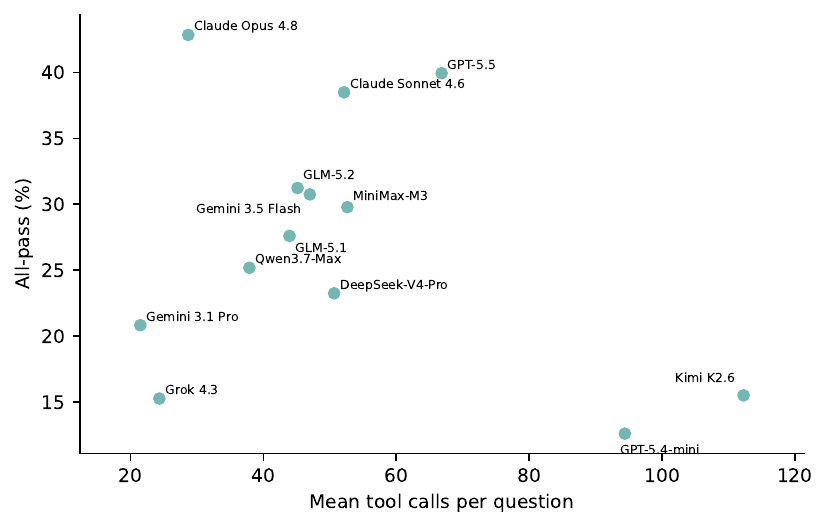}
\caption{All-pass against mean tool calls per question.}
\label{fig:effort-tools}
\end{figure}

\paragraph{Tool calls and correctness.}
Figure~\ref{fig:effort-tools} plots all-pass against mean tool calls per question,
as mentioned in Section~\ref{sec:effort}. Volume does not track correctness:
Claude Opus 4.8 leads the benchmark on 28.7 calls per question, while Kimi K2.6 issues
112.3 and GPT-5.4-mini 94.4.

\begin{figure}[h]
\centering
\includegraphics[width=0.8\linewidth]{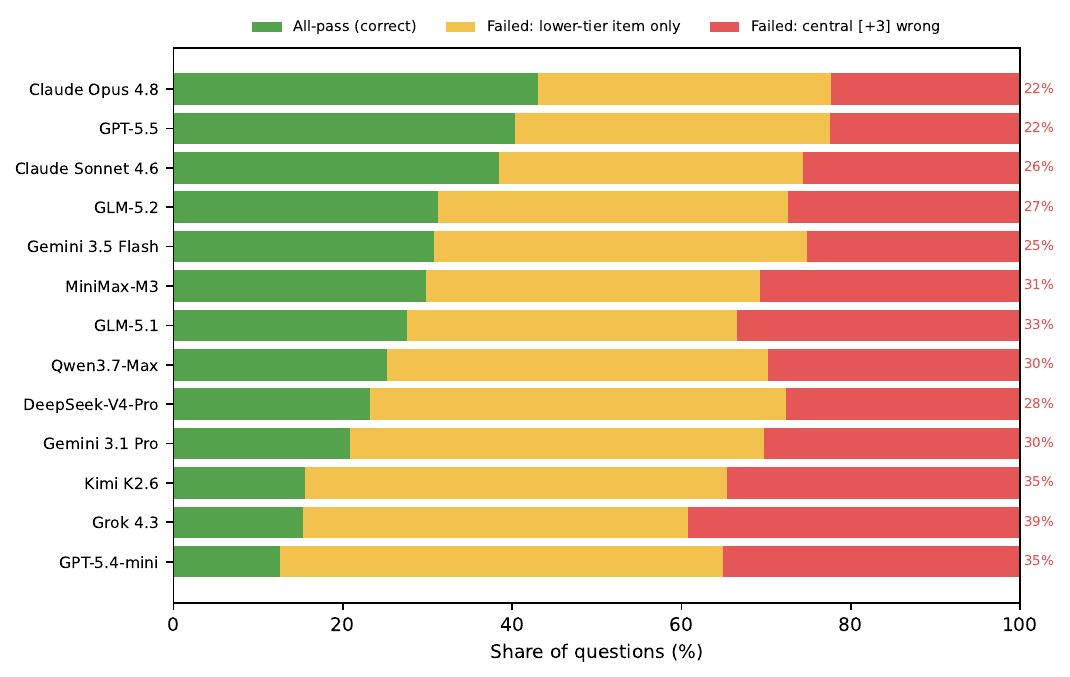}
\caption{Failure decomposition by model. Each evaluated question is correct
(all-pass), fails only on lower-tier [+1]/[+2] items, or is central-wrong (misses
a high-value [+3] item). Central errors (right, red) make up a large share of
every model's failures; the percentage of all questions that are central-wrong is
annotated.}
\label{fig:failures}
\end{figure}

\paragraph{Failure decomposition.}
Figure~\ref{fig:failures} classifies each evaluated question as all-pass, a
lower-tier-only failure (errs solely on [+1]/[+2] items), or central-wrong (misses
a high-value [+3] item). Central errors make up a large share of every model's
failures, accounting for roughly 41\% of all failures across the corpus.

\section{Answer length and verbosity}
\label{app:verbosity}
\begin{figure}[h]
\centering
\includegraphics[width=0.72\linewidth]{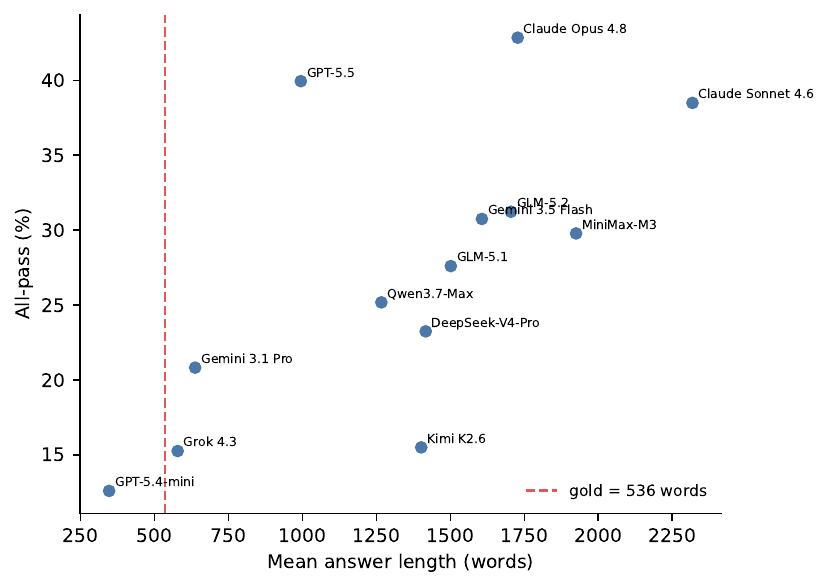}
\caption{Mean answer length versus all-pass per model, with the 536-word gold
reference marked. Answer length is positively associated with all-pass, yet every agent at the top of the leaderboard is several times more verbose than the expert references.}
\label{fig:answer-length}
\end{figure}

Most models produce answers far longer than the gold-standard references, which
average 536 words. Claude Sonnet 4.6 is the most verbose at roughly 2{,}320
words, more than four times the reference length, and only GPT-5.4-mini (about
350 words) falls below it. Answer length is positively associated with accuracy
across models (Spearman $\rho \approx 0.7$ with all-pass; Figure~\ref{fig:answer-length}),
though this appears to track underlying capability rather than verbosity itself:
GPT-5.5 attains the second-highest all-pass with one of the shortest answers
(about 990 words). The sharper contrast is with the expert references, which
reach a complete and correct answer in far fewer words than any strong agent. A
reading of the public-set answers indicates that this extra length is largely
structural and expository rather than substantive: agents organize their answers
with section headers and an issue-rule-application-conclusion structure, walk
through each statutory element explicitly even when it is uncontested, and recite
background frameworks and citation detail that the question does not require,
whereas the expert references state the holding and the controlling authorities
directly.

\section{Agent trajectories}
\label{app:traj}
\begin{figure}[t]
\centering
\includegraphics[width=\linewidth]{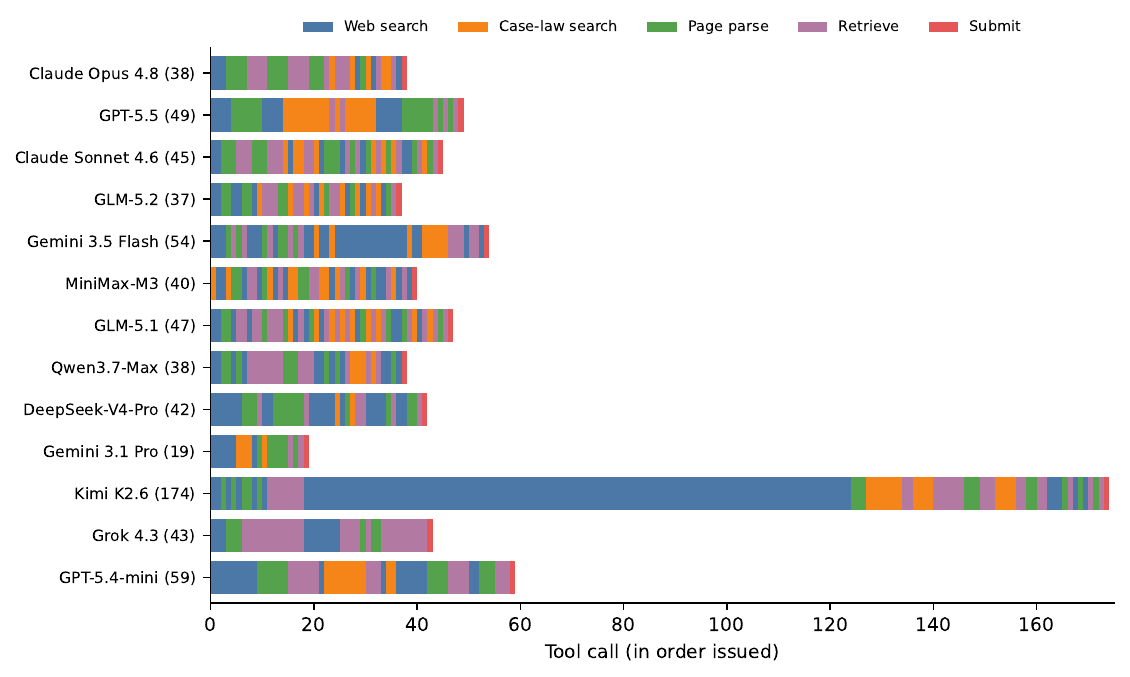}
\caption{Per-model tool-call trajectories on the sample question P-005, each call
bucketed by type and shown in the order it was issued; bar length is the number
of tool calls. Trajectory length varies more than tenfold and does not track
all-pass performance (models are ordered top-to-bottom by corpus-wide all-pass).}
\label{fig:trajectory}
\end{figure}

To illustrate how models allocate effort, we trace the ordered sequence of tool
calls on one question, P-005 (the Virginia equipment-leasing problem under UCC Article
2A presented in Appendix~\ref{app:sample}), bucketing each call by type
(Figure~\ref{fig:trajectory}). Trajectory length varies from 19 calls (Gemini 3.1 Pro)
to 174 (Kimi K2.6), while the five models that get a full score on the question cluster between 37 and 45 calls. Claude Opus 4.8 answers in 38 calls: web
and case-law search to frame the issue, CourtListener to read the controlling provisions
and decisions, page reading and retrieval to pull specific passages, then a single
submission. It cites eight authorities, all of which verify, but misses one rubric
criterion. Kimi K2.6 issues 174 calls, 117 of them web searches, and satisfies
every criterion, but cites no authority and so fails the source check. Grok 4.3 exposes a different behavior: it never queries CourtListener, working from web search alone, and cites three authorities for 76\% of the rubric weight. On this question
the three fail for distinct reasons: one grounded but incomplete, the other complete but
ungrounded, the third never reaching primary authority.

\end{document}